\documentclass[letterpaper]{article} 
\usepackage{aaai2026}  
\nocopyright
\usepackage{times}  
\usepackage{helvet}  
\usepackage{courier}  
\usepackage[hyphens]{url}  
\usepackage{graphicx} 
\usepackage{marvosym}
\usepackage{natbib}  
\usepackage{caption} 
\usepackage{bibentry}
\usepackage{algorithm}
\usepackage{algorithmic}
\usepackage{listings}
\usepackage{siunitx}
\usepackage{bibentry}
\usepackage{amsfonts}
\usepackage{amsmath}
\usepackage{amssymb}
\usepackage{hyperref}
\usepackage{physics}
\usepackage{listings}

\usepackage{float}
\usepackage{newfloat}
\usepackage{listings}
\DeclareCaptionStyle{ruled}{labelfont=normalfont,labelsep=colon,strut=off} 
\floatstyle{ruled}
\newfloat{listing}{tb}{lst}{}
\floatname{listing}{Listing}
\title{Rediscovering the Lunar Equation of the Centre with AI Feynman via Embedded Physical Biases}
\author {
    
    Saumya Shah,
    Zi-Yu Khoo,
    Abel Yang,
    St\'ephane Bressan\footnote{Deceased}
}
\affiliations {
    National University of Singapore\\
    Corresponding Author: saumya.shah@u.nus.edu
}

\begin{document}

\maketitle

\begin{abstract}
    This work explores using the physics-inspired AI Feynman symbolic regression algorithm to automatically rediscover a fundamental equation in astronomy -- the Equation of the Centre. Through the introduction of observational and inductive biases corresponding to the physical nature of the system through data preprocessing and search space restriction, AI Feynman was successful in recovering the first-order analytical form of this equation from lunar ephemerides data. However, this manual approach highlights a key limitation in its reliance on expert-driven coordinate system selection. We therefore propose an automated preprocessing extension to find the canonical coordinate system. Results demonstrate that targeted domain knowledge embedding enables symbolic regression to rediscover physical laws, but also highlight further challenges in constraining symbolic regression to derive physics equations when leveraging domain knowledge through tailored biases.
\end{abstract}

\section{Introduction}

A fundamental challenge in computational physics is the automated discovery of governing equations from observational data. While researchers have historically relied on intuition and theoretical frameworks to derive physical laws, recent advances in symbolic regression offer the potential to systematically extract mathematical relationships directly from measurements. However, the vast combinatorial space of possible equations makes brute-force approaches computationally infeasible, particularly for complex astrophysical systems where multiple interacting effects and observational noise obscure underlying principles \cite{karniadakisPhysicsinformedMachineLearning2021}. \emph{AI Feynman} is a physics-inspired symbolic regression algorithm developed by Silviu-Marian Udrescu and Max Tegmark in 2020 \cite{udrescu2020aifeynmanphysicsinspiredmethod} shown to be capable of rediscovering one hundred equations from the \emph{Feynman Lectures on Physics}. We investigate whether this algorithm can rediscover the Equation of the Centre from lunar ephemerides data. The Equation of the Centre quantifies the angular discrepancy between idealised uniform circular motion and actual elliptical Keplerian orbits, expressed as a series expansion in orbital eccentricity. Recovering this equation automatically would demonstrate that symbolic regression can navigate the complexity of real astronomical systems, including the challenge of distinguishing fundamental orbital dynamics from perturbative effects such as evection, variation, and solar gravitational interference.

Recent demonstrations have shown that AI Feynman can rediscover Feynman Lecture equations from synthetic data and, with manually embedded domain knowledge, recover the orbital equation of Mars from historical tables \cite{Khoo2023.1}. However, the Moon presents a substantially more difficult challenge due to its motion being dominated by numerous perturbations with magnitudes comparable to higher-order terms in the governing equations. Furthermore, symbolic regression algorithms typically lack the domain expertise to select appropriate coordinate systems, often converging on non-parsimonious solutions that obscure physical interpretations.

To address these challenges, we first systematically embed observational and inductive biases into AI Feynman by preprocessing lunar data into planar coordinates via principal component analysis, isolating anomalistic cycles to enforce periodicity, and restricting the function search space to trigonometric forms. Based on the limitations of this manual approach, we then propose a detailed framework for an automated preprocessing extension that transforms astronomical datasets into multiple reference frames relative to system bodies and barycentres, performs dimensionality reduction, and presents candidate equations along a combined Pareto frontier ordered by accuracy and parsimony. This proposed extension is designed to reduce the need for manual coordinate system selection while enabling AI Feynman to identify equations in canonical coordinate systems more reliably.
\section{Background}

\subsection{Astronomical Coordinate Systems}

Various coordinate systems are used to record positions of celestial bodies in astronomical research, and are specified by their origin and their basis. Angular coordinate systems define the celestial position of a body by the relative angle between a reference point, reference plane(s), and the body. Rectilinear or Cartesian coordinate systems define celestial positions using a reference point (origin) and three coordinate axes corresponding to the three spatial dimensions. Common origin points include the barycentre of the solar system or the galaxy. \cite{Karttunen2016-gt}

\subsubsection{Canonical Coordinates}
A canonical coordinate system is defined as a coordinate system in which the equations governing the dynamics of a system are expressed in a standard, simplified form that minimises computational and symbolic complexity.

One common canonical coordinate system in astronomy is the barycentric coordinate system, where the origin is set at the centre of mass (barycentre) of a gravitationally interacting system, such as a planetary system. This system is advantageous because it allows for simplification of the equations governing orbital motion by reducing the relative accelerations and gravitational interactions to those acting directly on each mass from this central point. 

\subsection{Equation of the Centre} \label{EqnOfCentreSection}
In Keplerian motion, a body in orbit around another follows an elliptical, periodic path. The Equation of the Centre describes the angular difference between a uniform circular orbit and a Kepler elliptical orbit. More formally, it refers to the difference between the \textit{mean anomaly} corresponding to the angular distance in a uniform circular orbit, and the \textit{true anomaly} corresponding to the angular distance in a Keplerian orbit.

This inequality, caused by the elliptical shape of the moon's orbit, may be expressed as a series using Bessel functions of the first kind as a function of eccentricity, as shown in equation \eqref{eqnofcentre} \cite{brown1896introductory}:

\begin{equation} \label{eqnofcentre}
\resizebox{0.90 \columnwidth}{!}{$
v - M = \\2 \sum_{s = 1}^{\infty}\frac{1}{s} \left\{ J_s(se) + \sum_{p = 1}^{\infty} \beta^{p} (J_{s-p}(se) + J_{s+p}(se)) \right\} \sin{sM}
$}
\end{equation}

This can be simplified to equation \eqref{eqnofcentresimplified}\\
\begin{equation}\label{eqnofcentresimplified}
v - M = \left(2e - \frac{1}{4}e^3 + \frac{5}{96}e^5 + \frac{107}{4608}e^7\right) \sin{M}  + ...
\end{equation}

Where
\begin{align*}
    v & \text{ is the true anomaly}\\
    M & \text{ is the mean anomaly}\\
    J_n & \text{ are the Bessel functions} \\
    \beta &= \frac{1}{e}\left(1 - \sqrt{1 - e^2}\right) \\
    e & \text{ is the eccentricity of the orbit}
\end{align*}

\subsection{Lunar Perturbations}\label{sec:lunpert}

The Moon follows a Keplerian elliptical orbit. Lunar perturbations are deviations from a uniform circular orbit of the Moon around Earth. They can be further classified into \emph{inequalities}, which are specific periodic lunar perturbations that affect the right ascension of the Moon, the celestial equivalent of the longitude. The Equation of the Centre is an example of a lunar inequality.\\
Other prominent lunar inequalities include \emph{evection} (an approximately monthly variation in the eccentricity and right ascension of the lunar perigee) and \emph{variation} (semi-monthly lunar acceleration) \cite{brown1896introductory}.

\subsubsection{Causes of Perturbations} 

There are three main causes of these perturbations:

\emph{Apsidal Precession} is the gradual rotation of the apsidal line, which is the major axis of orbit from the perigee to the apogee. This changes the timing of the perigee and apogee with an 8.85 year cycle \cite{2024amos.conf..128R}. It does not change the overall shape of the orbit.

\emph{Nodal precession} is the gradual rotation of the orbital plane around the rotational axis caused by Earth's oblateness. It changes the orientation of the lunar orbital plane with the ecliptic with a cycle of 18.6 years \cite{deBoer1993}. It also leads to variations in the strength of the tides. This has a minor effect on the major axis of the orbit, but does not change the overall shape of the orbit.

\emph{Gravitational tides} are the stretching of a body due to the gravitational force of another body. The sun's gravity stretches the Earth, which leads to a difference in gravitational pull on the moon over its orbit, which leads to an elliptical orbit \cite{ZACCAGNINO2020103179}. This also causes a small true libration effect.

\section{Related Work}
\subsection{Classical Symbolic Regression}
The main goal of symbolic regression is to find an analytic expression for an unknown function $f(\cdot)$ that maps the $d$-dimensional input $x \in \mathbb{R}^d$ to the target variable $y \cong f(x) \in \mathbb{R}$ given a dataset of observations $\{x_i, y_i\}_{i=1}^{N}$. However, finding equations that capture datasets is a combinatorial challenge; the sheer number of combinations of operators and operands makes a brute-force approach computationally unfeasible \cite{karniadakisPhysicsinformedMachineLearning2021}. There exist many techniques for symbolic regression, but they can be broadly divided into three main classes: expression tree-based, regression-based, and physics- or mathematics-informed \cite{makke2023interpretablescientificdiscoverysymbolic}.

Expression tree-based methods are often based on paradigms like genetic programming, where models can discover the form and coefficients of the equation by representing approximate candidate solutions using an expression tree-like data structure. Transition functions, like random recombination or permutation, are iteratively applied to generate new candidate solutions, while candidate solutions with low `fitness' - some desired objective function - are dropped from the model \cite{oh2023geneticprogrammingbasedsymbolic}.

Regression-based methods, on the other hand, search for the coefficients of a fixed prespecified basis that minimise error. As the size of the basis increases, the accuracy of the function may increase, but the form of the solution may grow less parsimonious \cite{makke2023interpretablescientificdiscoverysymbolic}.

\subsection{Physics-Informed Symbolic Regression}
Physics-informed symbolic regression methods leverage simplifying properties derived from physics, like symmetry and separability to limit the search space and find parsimonious and accurate solutions along the Pareto frontier, which represents the solutions with the best trade-offs between parsimony and accuracy of the solution to the system in question. The introduction of simplifying physical properties generally takes three forms of biases \cite{Khoo2023.1}: observational bias, learning bias, and inductive bias. Observational biases are introduced through the selection of data augmentation and transformation techniques for the data to embody underlying physical principles \cite{karniadakisPhysicsinformedMachineLearning2021}. Learning biases include the choice of appropriate loss functions, hyperparameters, and learning algorithms that guide the model toward physically meaningful solutions. Inductive biases are the inherent assumptions built into the architecture of the model such that predicted solutions are guaranteed to satisfy a set of physical conditions and laws. There are various techniques that have been shown to be effective for physics-informed symbolic regression.

\subsubsection{SINDy}\label{sec:sindy}
\emph{Sparse Identification of Non-Linear Dynamics} \cite{Brunton2016} leverages the sparsity of key terms in physical systems, using sparsity techniques for efficient identification of relevant terms in the model. This promotes parsimony and avoids overfitting. The method involves collecting state data and its derivatives (possibly approximated numerically), adding noise for robustness, and constructing a library of candidate non-linear functions for each state variable. A sparse regression technique, like LASSO, is then applied to determine the coefficients that identify the important terms within the model. Domain knowledge can further guide the selection of non-linear functions, and help exploit other simplifying properties. SINDy has been shown to be effective in recovering accurate models for chaotic systems like the Lorenz system and vortex shedding, demonstrating robustness to noise and even the absence of direct derivative measurements. However, challenges remain in choosing the most suitable measurement coordinates and the optimal basis of the sparsifying function.

\subsubsection{Graph Neural Networks \& PySR}\label{sec:gnn}
\emph{Lemos et al} \cite{Lemos2023} demonstrate the utility of embedding inductive biases in rediscovering Newton's Law of Gravitation from trajectory data of solar system objects. First, a graph neural network is used to simulate the dynamics of solar system objects from 30 years of trajectory data, with the positions and velocities of the bodies represented as nodes, and physical interactions as edges between these nodes. Inductive biases, such as translational invariance, rotational invariance, and Newton's laws of motion were embedded through data augmentation and the multiplicative relationship between the node and its acceleration. This promoted candidate solutions that were aligned with existing known physical laws. Then, an open-source analogue of Eureqa (implemented in the \lstinline{PySR} library \cite{cranmer2023interpretablemachinelearningscience}) was used to discover analytical expressions for the learned simulator, where a tree search algorithm was used to produce a set of candidate functions, which were evaluated using a score corresponding to the ratio between accuracy and parsimony. This two-step method has been shown to be effective and efficient in discovering analytic equations corresponding to Newton's Law of Gravitation. The authors identify the implementation of the method using Bayesian Neural Networks to model the masses in the system as an avenue for future exploration. The authors further identify the evaluation score for the candidate solutions as a limitation, emphasising that it may not align with what a physicist may identify as a `good' equation.

\subsubsection{AI Feynman}
\emph{AI Feynman} \cite{udrescu2020aifeynmanphysicsinspiredmethod} utilises neural networks to identify simplifying physical properties within the data. This approach addresses the limitations of techniques like genetic algorithms and sparse regression, which might struggle to capture these underlying principles. In this regard, AI Feynman outperforms the techniques discussed in previous sections. It incorporates six assumptions about the underlying function, including known physical units of variables, low-order polynomial structures, smoothness, composition, symmetry, and separability. The core algorithm works recursively, first employing dimensional analysis to reduce data complexity and then fitting polynomials and exploring increasingly complex expressions through brute force. Additionally, AI Feynman uses neural networks to identify specific transformations like symmetry, separability, and variable equality, allowing for a more efficient decomposition of the problem into simpler sub-problems with fewer variables. This focus on decomposability is a key improvement over methods like Eureqa. Khoo et al. have demonstrated the effectiveness of AI Feynman with embedded observational and inductive biases in recovering the orbital equation of Mars from the Rudolphine tables \cite{Khoo2023.1}.

Despite progress, automated symbolic regression remains limited by its reliance on expert-driven preprocessing (especially choice of coordinate system) and challenges in distinguishing fundamental physical effects from observational noise and perturbations. These open problems motivate work on automated, physics-guided transformations that can steer equation discovery toward canonical, interpretable forms.

\section{Methodology}
\subsection{Overview}
This section introduces the experimental setup, including the dataset and its preprocessing and augmentation, and the techniques used to introduce observational and inductive bias to help AI Feynman rediscover the Equation of the Centre (\eqref{eqnofcentre} and \eqref{eqnofcentresimplified}) from lunar orbital data. The experiments were subsequently evaluated on their ability to recover these equations.

\subsection{Dataset}
Geocentric lunar ephemeris data from between 2024-01-01 00:00:00 and 2025-01-01 00:00:00 with a step size of 60 minutes was obtained from NASA JPL's Horizons System \cite{horizons}. The relevant features of the dataset include the datetime, right ascension (in hours-minutes-seconds of time), declination (in degrees-minutes-seconds of arc), and delta (the geocentric distance to the Moon in AU)\\

\subsection{Preprocessing}
The equatorial coordinates (right ascension and declination) were converted to radians, and the ecliptic coordinates (longitude and latitude) were calculated in radians using the Right Ascension and Declination values and AstroPy's in-built \lstinline{transform_to} method \cite{astropy:2013, astropy:2018, astropy:2022}.

For both the coordinate systems, principal component analysis was performed using the \lstinline{sklearn} \cite{scikit-learn} library to obtain equatorial and ecliptic planar coordinates over the span of the dataset. This is an observational bias where the planarity of the lunar orbit is introduced directly through the choice of reference frame.

The data for the individual anomalistic (apogee to apogee) lunar cycles were isolated by determining the timestamps of the maxima in the planar coordinate values, and assigning the data points corresponding cycle numbers, for a total of 14 unique cycles. This is an observational bias where the periodicity of the lunar orbit is used to simplify the dataset.

Finally, the true anomaly, mean anomaly, and residuals were calculated for the moon for each lunar cycle using the planar coordinates. The mean anomalies were calculated by finding the time since the perigee for each point in the cycle and normalising the time since the perigee by the total time of the cycle. The true anomalies were calculated iteratively using Kepler's first and second laws \cite{brown1896introductory}. The circular motion residuals, which correspond to the difference between an idealised uniform circular orbit and a Keplerian orbit, were calculated by subtracting the mean anomalies from the true anomalies.

\begin{figure}
    \centering
    \includegraphics[width=\linewidth]{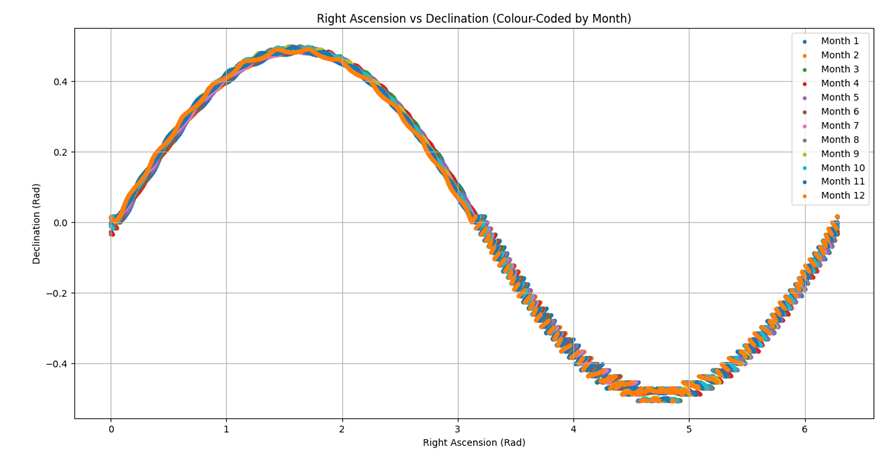}
    \caption{Monthly Lunar Declination vs Right Ascension}
    \label{fig:placeholder}
\end{figure}

\subsection{Experimental Setup}\label{sec:expSetupLunar}

We conduct three experiments with different combinations of inputs for AI Feynman, corresponding to the inclusion of different observational biases and inductive bases as shown in Table \ref{tab:vartablelunar}.

\begin{table}[H]
    \centering
    \resizebox{\columnwidth}{!}{%
    \begin{tabular}{|c|c|c|c|}\hline
         &\textbf{Input Variable(s)} & \textbf{Target Variable} & \textbf{Search Space Bias} \\\hline
         1 & $M$ & Circular motion residuals & All Functions \\\hline
         2 & $M$ & Circular motion residuals & Trigonometric Functions \\\hline
         3 & $\sin{(M)}, \sin{(2M)}, \sin{(3M)} $ & Circular motion residuals & Trigonometric Functions \\\hline
    \end{tabular}%
    }
    \caption{Input \& Target Variables for AI Feynman for Lunar Orbit ($M$ corresponds to the mean anomaly)}
    \label{tab:vartablelunar}
\end{table}

The first experiment applies AI Feynman directly without biases. The second experiment is inductively biased where the increased bias of AI Feynman's search space towards trigonometric functions embodies the periodic nature of the orbit. The third experiment also directly embeds an observational bias corresponding to the periodicity of the orbit through the augmentation of the dataset. Thus, the modifications to embed observational biases inform AI Feynman of the periodicity of the lunar orbit and the trigonometric nature of the mean anomaly, replacing it with the sine of mean anomaly.

The measure of fit places a logarithm-scaled penalty on the absolute loss.
The measure of parsimony places a logarithm-scaled penalty on real numbers,
variables and operators in an equation. These measures are used to compare solutions generated along the Pareto frontier for the experiments described in Table \ref{tab:vartablelunar}.

\section{Performance Evaluation}

The solutions along the Pareto frontier are reported in column 2 of tables \ref{tab:lunarExp1}, \ref{tab:lunarExp2}, and \ref{tab:lunarExp3}. $M$ indicates the mean anomaly. All solutions were simplified; those that evaluated to constants independent of the input variable were discarded.\\

\textbf{Experiment 1}
\renewcommand{\arraystretch}{3}
\begin{table}[H]
    \centering
    \resizebox{\columnwidth}{!}{%
    \begin{tabular}{|c|c|c|c|}\hline
         &\textbf{Equation} & \textbf{Measure of Fit} & \textbf{Measure of Parsimony} \\\hline
         1a & $\frac{M}{M^2+M+1}-0.2$ & 24.0474 & 17.9316 \\\hline
         1b & $\frac{M}{M^2+2}-0.22$ & 23.0438 & 20.5344 \\\hline
         1c & $\frac{2M}{M^2+2}-0.877$ & 26.091250 & 23.02309 \\\hline
    \end{tabular}
    }
    \caption{AI Feynman Solutions for Lunar Experiment 1}
    \label{tab:lunarExp1}
\end{table}

\textbf{Experiment 2}
\renewcommand{\arraystretch}{3}
\begin{table}[H]
    \centering
    \resizebox{\columnwidth}{!}{%
    \begin{tabular}{|c|c|c|c|}\hline
         &\textbf{Equation} & \textbf{Measure of Fit} & \textbf{Measure of Parsimony} \\\hline
         2a & $-\arctan(0.2M)$ & 26.4463 & 15.9315 \\\hline
         2b & $-0.4M$ & 26.0228 & 11.3219 \\\hline
    \end{tabular}
    }
    \caption{AI Feynman Solutions for Lunar Experiment 2}
    \label{tab:lunarExp2}
\end{table}

\textbf{Experiment 3}
\renewcommand{\arraystretch}{3}
\begin{table}[H]
    \centering
    \resizebox{\columnwidth}{!}{%
    \begin{tabular}{|c|c|c|c|}\hline
         &\textbf{Equation} & \textbf{Measure of Fit} & \textbf{Measure of Parsimony} \\\hline
         3a & $0.1095\sin M$ & 25.8333 & 4.3114 \\\hline
         3b & $0.1142857\sin(M)$ & 25.3271 & 7.9773 \\\hline
         3c & $0.1146627\sin(M)$ & 25.3483 & 7.9921 \\\hline
         3d & $0.52524 \sin(M)\left(\sqrt{\sqrt{\sin(M)+2}}+1\right)$ & 25.2808 & 62.7726 \\\hline
    \end{tabular}%
    }
    \caption{AI Feynman Solutions for Lunar Experiment 3}
    \label{tab:lunarExp3}
\end{table}

We observe that none of the equations in experiments 1 and 2 match the expected form of the Equation of the Centre, neither \eqref{eqnofcentre} nor \eqref{eqnofcentresimplified}. In experiment 3, we note that when the observational bias of the trigonometric nature of the mean anomaly is embedded, we observe equations symbolically similar to the first-order term of the equation \eqref{eqnofcentresimplified}. We observe that by comparing the coefficient of $\sin(M)$ in $\eqref{eqnofcentresimplified}$ and 3A - 3C, we obtain estimates of 0.0547705, 0.057354933682, and 0.0571662 for the eccentricity of the lunar orbit, all of which are remarkably close to the average observed lunar eccentricity of 0.0549 \cite{brown1896introductory}, with equation 3A having the least deviation of $0.235$ per cent between the true value of eccentricity and the eccentricity estimate obtained from AI Feynman. Thus, equation 3A corresponds to the first-order form of the Equation of the Centre \eqref{eqnofcentresimplified}.

\section{Discussion and Further Work}

In this work, we applied the AI Feynman symbolic regression algorithm in an attempt to rediscover a fundamental equation governing lunar motion from observational data. The introduction of various observational and inductive biases corresponding to the periodicity, planarity, and trigonometric nature of the system, such as planar coordinates, restrictions to anomalistic cycles, conversion of angular inputs to their sines, and an increased bias towards the trigonometric function space were sufficient in constraining the search space such that AI Feynman was able to recover the expected first-order form of the Equation of the Centre from lunar ephemerides data, something that was previously possible only with human intuition and physical understanding.

A limitation of AI Feynman was its inability to discover higher-order terms of the Equation of the Centre \eqref{eqnofcentre}. This is likely because the magnitudes of higher-order terms of the Equation of the Centre are much smaller, similar in magnitude to other lunar inequalities (as described in the section on lunar perturbations). This makes it difficult for AI Feynman to fit the higher-order terms of the Equation of the Centre without decomposing the other lunar inequality components.

Another limitation of AI Feynman was its inability to infer the canonical coordinate system from the data, resulting in physically accurate candidate solutions being dominated by more parsimonious but less accurate solutions due to the extra computation required to transform the coordinates to their canonical system. To address this critical limitation of canonical coordinate inference, we propose an automated preprocessing extension as a framework for our future work.

This extension would be designed to autonomously generate and evaluate multiple, physically relevant reference frames. The core principle is to create an automated process that transforms observational data, such as $N$-body trajectories, into several coordinate systems relative to system bodies and their barycentres

For each transformed dataset, the framework would apply dimensionality reduction techniques and subsequently execute the symbolic regression algorithm to find candidate equations for a specified target variable. By aggregating the candidate equations from all tested reference frames into a unified Pareto frontier, the system could identify the solution that offers the optimal trade-off between accuracy and parsimony. It is hypothesised that the most parsimonious, physically correct equation, representing the canonical form of the governing law, would naturally emerge from this process.

Validating this framework would necessitate a synthetic dataset where the underlying dynamics and ground-truth equations are known \textit{a priori}, such as a simulated $N$-body system. This would allow for a rigorous quantitative assessment of the extension's ability to correctly identify the canonical coordinate system from the set of all tested frames.

\section{Conclusion}

This work demonstrates that AI Feynman can successfully rediscover fundamental astronomical equations from observational data when appropriately guided by domain knowledge. Through systematic embedding of observational and inductive biases, including planar coordinate transformation via PCA, isolation of anomalistic cycles, and restriction to trigonometric function spaces, AI Feynman recovered the first-order analytical form of the lunar Equation of the Centre from geocentric lunar ephemerides data. The derived eccentricity estimate exhibited very low deviation from the established lunar eccentricity value, validating the physical accuracy of the symbolic regression approach.

However, this success required substantial manual preprocessing and expert-driven coordinate system selection, revealing critical limitations in current automated equation discovery methods. AI Feynman's inability to infer canonical coordinate systems autonomously resulted in physically meaningful solutions being obscured by more parsimonious alternatives expressed in suboptimal reference frames. Furthermore, the algorithm failed to capture higher-order terms of the Equation of the Centre, as their magnitudes remained comparable to perturbative effects from other lunar inequalities such as evection and variation, preventing effective decomposition of the fundamental orbital dynamics from observational noise.

To address these limitations, we proposed a comprehensive automated preprocessing framework which aims to eliminate the dependency on manual coordinate selection by systematically transforming astronomical datasets into multiple reference frames to construct a unified Pareto frontier of candidate equations across all coordinate systems. This approach provides a way toward more robust, generalisable symbolic regression by enabling the algorithm to autonomously identify the canonical representation of a system's governing equations.

Future work must focus on developing more sophisticated bias embedding techniques that can isolate higher-order terms in the presence of comparable perturbative effects, potentially through hierarchical decomposition strategies or multi-scale temporal analysis. Additionally, the proposed automated coordinate finding extension requires empirical validation on synthetic datasets and subsequent testing on complex astronomical systems beyond the Earth-Moon system. Extending this framework to incorporate additional physical constraints, such as conservation laws, symmetry principles, and known scaling relationships, may further enhance the capability of symbolic regression algorithms to autonomously discover governing equations in canonical forms across diverse scientific domains. The integration of such automated, physics-guided preprocessing with existing symbolic regression methods represents a critical step toward realising the vision of computational physics as a tool for genuine scientific discovery rather than merely sophisticated curve fitting.

\section{Acknowledgements}
Professor St\'ephane Bressan, who passed away before this work's publication, is remembered for his foundational guidance and mentorship.\\

\textbf{Data and Materials Availability}: All code used for this paper is available at \url{https://github.com/lordsaumya/AI-Feynman}. The lunar ephemerides dataset is available at \url{https://ssd.jpl.nasa.gov/horizons/}. Analysis details are available upon request.

\newpage
\nocite{*}
\bibliography{sources}

\end{document}